\documentclass[letterpaper, 10pt, conference]{IEEEtran}

\IEEEoverridecommandlockouts       

\usepackage{endnotes}
\usepackage{amsmath,amssymb}
\usepackage{graphicx}
\usepackage{subcaption}
\usepackage{color}
\usepackage{algorithm}
\usepackage[noend]{algpseudocode}
\usepackage[utf8]{inputenc}
\usepackage{tabularx}
\usepackage{tabulary}
\usepackage{multirow}
\usepackage{setspace}
\usepackage{ragged2e}
\usepackage{lscape}
\usepackage{float}
\usepackage{gensymb}
\usepackage[table,xcdraw]{xcolor}
\usepackage{placeins}
\usepackage{float}
\usepackage{flushend}
\usepackage{dblfloatfix}

\newcolumntype{w}[1]{>{\raggedleft\hspace{0pt}}p{#1}}
\newcolumntype{x}[1]{>{\centering\hspace{0pt}}p{#1}}

\newcolumntype{w}[1]{>{\raggedleft\hspace{0pt}}p{#1}}
\newcolumntype{x}[1]{>{\centering\hspace{0pt}}p{#1}}

\newcommand\MyBox[2]{
  \fbox{\lower0.75cm
    \vbox to 1.7cm{\vfil
      \hbox to 1.7cm{\hfil\parbox{1.4cm}{#1\\#2}\hfil}
      \vfil}%
  }%
}

\newlength{\Oldarrayrulewidth}

\begin{document}
\title{Railway LiDAR semantic segmentation based on intelligent semi-automated data annotation}

\author{\IEEEauthorblockN{Florian Wulff, Bernd Schäufele}
\IEEEauthorblockA{\textit{Fraunhofer Institute for Open } \\
\textit{Communication Technologies (FOKUS)}\\
Berlin, Germany \\
\{florian.wulff, bernd.schaeufele\}@fokus.fraunhofer.de}
\and
\IEEEauthorblockN{Julian Pfeifer, Ilja Radusch}
\IEEEauthorblockA{\textit{Daimler Center for Automotive} \\
\textit{Information Technology Innovations (DCAITI)}\\
Berlin, Germany \\
\{julian.pfeifer, ilja.radusch\}@dcaiti.com}
}

\maketitle

\begin{abstract}
Automated vehicles rely on an accurate and robust perception of the environment. Similarly to automated cars, highly automated trains require an environmental perception. Although there is a lot of research based on either camera or LiDAR sensors in the automotive domain, very few contributions for this task exist yet for automated trains. Additionally, no public dataset or described approach for a 3D LiDAR semantic segmentation in the railway environment exists yet. Thus, we propose an approach for a point-wise 3D semantic segmentation based on the 2DPass network architecture using scans and images jointly. In addition, we present a semi-automated intelligent data annotation approach, which we use to efficiently and accurately label the required dataset recorded on a railway track in Germany. To improve performance despite a still small number of labeled scans, we apply an active learning approach to intelligently select scans for the training dataset.
Our contributions are threefold: We annotate rail data including camera and LiDAR data from the railway environment, transfer label the raw LiDAR point clouds using an image segmentation network, and train a state-of-the-art 3D LiDAR semantic segmentation network efficiently leveraging active learning. The trained network achieves good segmentation results with a mean IoU of 71.48\% of 9 classes.
\end{abstract}

\section{Introduction}
\label{sec:introduction}

The field of automated driving experiences remarkable progress and aims to disrupt the transportation industry and individual mobility. In parallel to the research on automated cars, the automation of trains also constantly increases. However, in comparison to the automotive sector, the environmental perception of highly automated trains sees relatively little public research. To support and assist the train driver and to ensure safe and reliable reactions to special circumstances, e.g., by obstacle detection and emergency braking systems, a robust and well-trained perception of the surroundings is necessary even under challenging conditions. Here, LiDAR is a capable sensor to robustly capture highly accurate three-dimensional data. Therefore, we propose a LiDAR semantic segmentation approach for environmental perception.

The three-step approach involves dataset preparation, dataset labeling, and model training. First, the recorded LiDAR data is filtered and cropped to remove noise, outliers, and reflections. The LiDAR scans are then synchronized to the respective camera image and motion distortion is corrected. Secondly, a dataset consisting of 602 coarsely annotated LiDAR scans is created. To achieve this with minimal manual effort, the image semantic segmentation network Deeplabv3+ \cite{deeplabv3} is trained using the Railsem19 \cite{railsem19} dataset. The predicted 2D labels are mapped to the closest 3D LiDAR point. Afterwards, some wrongly labeled points that result from calibration inaccuracies, are corrected manually. Thereby, a small, precisely labeled dataset of 52 scans is created. In the third step, 2DPass \cite{2dpass}, a state-of-the-art LiDAR semantic segmentation network, is trained. Finally, using an active learning approach, we select and label additional scans to improve the segmentation performance iteratively. With this, very good segmentation results with a mean IoU of 71.48\% of 9 classes are achieved.

The paper is structured as follows: In section \ref{sec:related}, we introduce related work and datasets. In section \ref{sec:data_preprocessing}, we describe the preprocessing and preparation of the raw data. In section \ref{sec:labeling}, we propose our highly efficient labeling approach to create the dataset. In section \ref{sec:training}, we explain the network architectures and the training process. Additionally, we improve the resulting performance by using active learning. In section \ref{sec:evaluation}, we evaluate and visualize the segmentation results. The paper concludes with a summary and future work in section \ref{sec:summary}.



\section{Related Work}
\label{sec:related}

Due to a lack of public datasets or benchmarks, and mostly corporate or confidential research, there are relatively few publications concerning perception in the railway environment.

\paragraph*{\textbf{Environmental perception for rail}}

Wang et al. \cite{wang2019railnet} and Katar and Duman \cite{katar2022automated} provide contributions for semantic segmentation of rail tracks or multiple object classes in images from an ego perspective, while Tong et al. \cite{tong2021fully} and Mammeri et al. \cite{mammeri2021uav} use convolutional neural networks to segment railway scenes from UAV aerial images.
Other rail-specific perception tasks covered in publications focus on perception tasks like signal detection \cite{marmo2006railway, liu2021real}, vegetation monitoring \cite{kafetzis2020uav, rahman2021vegetation} or catenary pole detection \cite{pastucha2016catenary, jung2016multi}. Advanced driver-assistant systems are another key area of rail research. As a contribution to those applications, Wang et al. \cite{wang2018efficient} and Ziegler et al. \cite{ziegler2023comprehensive} show the implementation of track and switch detection approaches by predicting splines or the track geometry, while Rodriguez et al. \cite{rodriguez2012obstacle} and Guan et al. \cite{guan2022lightweight} propose approaches for train and object detection. Mukojima et al. \cite{mukojima2016moving} and Etxeberria-Garcia et al. \cite{etxeberria2020application} focus on obstacle detection algorithms specifically for collision and free space detection systems.

\paragraph*{\textbf{Public datasets in the rail environment}}
In contrast to environmental perception in the automotive domain, where extensive and public datasets for semantic segmentation such as the SemanticKITTI dataset \cite{behley2019semantickitti}, the Waymo Open Dataset \cite{sun2020scalability}, or the nuScenes dataset \cite{caesar2020nuscenes} are available, there hardly exist any public datasets for environmental perception in the rail environment. This is due to the fact, that rail infrastructure is usually privately owned, or the access is highly restricted, and operating trains requires high effort and is very expensive.

Railsem19 \cite{railsem19} provides 8500 different camera images and complete high-quality semantic annotations for 22 classes, but no LiDAR data. Yet, for semantic segmentation for rail, this is the only suitable published dataset. The recently published OSDaR23 dataset \cite{osdar23} offers 1534 annotated frames including LiDAR and camera data with object annotations as bounding boxes, polygons and cuboids in 2D and 3D, but lacks semantic segmentation labels for environmental classes and point-wise labels for all points. Some point-wise semantic labels for objects only can be extracted from the 3D cuboids. Other datasets from the rail sector are either not public or provide only specific classes or labels for signal recognition tasks, such as FRSIGN \cite{frsign} or GERALD \cite{gerald}. After extensive research, no public general purpose dataset for semantic segmentation of LiDAR scans exists.

\paragraph*{\textbf{3D semantic segmentation}}
Currently, there is no related work on semantic scene segmentation for rail using deep neural networks with 3D scans from an onboard LiDAR sensor. 
Oude Elberink et al. \cite{oude2013rail} and M. Arastounia \cite{arastounia2015automated} segment individual parts of catenary poles and tracks in LiDAR scans but use traditional machine learning approaches.
Zhu and Hyyppa \cite{zhu2014use} extract railroad objects from very high-resolution airborne and mobile LiDAR scans and classify the railway environment in the 3D scans using statistical parameters and traditional machine learning approaches.

In contrast, semantic segmentation is well-established and very advanced in the automotive domain. For semantic segmentation in the image domain, DeepLabv3+ \cite{deeplabv3}, PSPNet \cite{zhao2017pyramid}, U-Net \cite{ronneberger2015u} or VGG19 \cite{simonyan2014very} are powerful and widely used networks. In the 3D LiDAR semantic segmentation benchmark, SphereFormer \cite{lai2023spherical}, RangeFormer \cite{kong2023rethinking}, 2DPass \cite{2dpass} or PVKD \cite{hou2022point} are examples of state-of-the-art networks for 3D semantic segmentation. We show, that these well-established networks can be applied to the rail environment.

\section{Sensor data preprocessing and preparation}
\label{sec:data_preprocessing}


\begin{figure*}[!htb]
 \centering
  \includegraphics[width=0.98\textwidth, trim=0 20 0 10, clip]{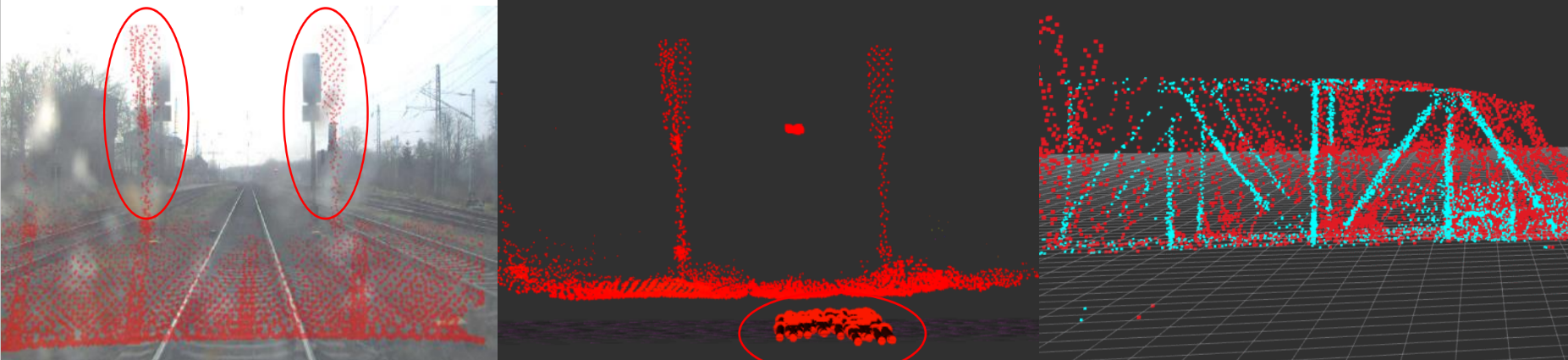} 
   \caption[corrected-lidar-scan]{Time offset (left), sensor noise and reflections (center), and correction of motion distortion (right) in the scans}
  \label{fig:scancorrection}
\end{figure*}

Due to the lack of existence of public 3D LiDAR railway datasets to train a semantic segmentation network for environmental perception, we create a rail-specific annotated dataset. We use a non-public dataset of a train run in the Ore Mountains region near Markersbach in Germany for our work. The point clouds cointain approximately 30,000 points and are captured at a rate of 4 Hz. The color images have a resolution of 2048x1536 at a rate of 1.5 Hz. Additionally, the 2D-3D transformations and camera parameters are provided.

As only raw data is available in the dataset, additional preprocessing steps are applied to produce a suitable dataset with high quality. First, the camera and LiDAR recordings are not synchronized. Timestamps of LiDAR scans and images have an offset of up to 300 ms in the worst case and 150 ms on average due to the different recording frequencies of both sensors. The offset can be seen in the left image in figure \ref{fig:scancorrection}. To synchronize the data, only scan pairs with less than 10 ms difference were selected. After time synchronization, approximately one-third of the total scans are remaining.

Secondly, noise, measurement errors, and reflections in the LiDAR point clouds are removed from the LiDAR data and filtered. The LiDAR is placed in a housing at the front of the train which is required for protection and to fulfill strict safety regulations in the rail sector. The housing causes reflections of the laser beams close to the origin as shown in the center image in figure \ref{fig:scancorrection}. After filtering the raw point cloud, about one-third of the recorded points remain. Thus, the resulting point cloud is very sparse with approximately 10,000 points per scan on average remaining.

Thirdly, the train moves at up to 100 kph, causing a movement of up to 2.8 m during 100 ms recording time for a single scan. This results in both a spatial offset between the image and the point cloud and a motion distortion of the objects in the point cloud.
Static structures are therefore not true to scale and not planar as shown in red in the right image in figure \ref{fig:scancorrection}. The distortion is corrected by calculating the movement with regard to the common origin induced by the motion for each point in the cloud and by shifting the point accordingly. This requires determining the offset corresponding to the distance traveled by current speed $\text{v}_{current}$ between the point timestamp $\text{t}_{point}$ and the initial timestamp of the scan $\text{t}_{scan}$ of the train for each point. The corrected point cloud is shown in turquoise in the right image in figure \ref{fig:scancorrection} based on the formula:
\begin{equation}
\text{offset}_{point} = (\text{t}_{point} - \text{t}_{scan}) \cdot \text{v}_{current}
\end{equation}

The created dataset has high quality and is filtered, calibrated, rectified, and synchronized. However, the total number of scans is reduced significantly, and the remaining scans are very sparse. The dataset contains 602 scans with a total of 652,000 points.

\section{Semi-automated dataset annotation}
\label{sec:labeling}

\begin{figure*}[h!tb]
 \centering
  \includegraphics[width=0.98\textwidth, trim=0 20 0 50, clip]{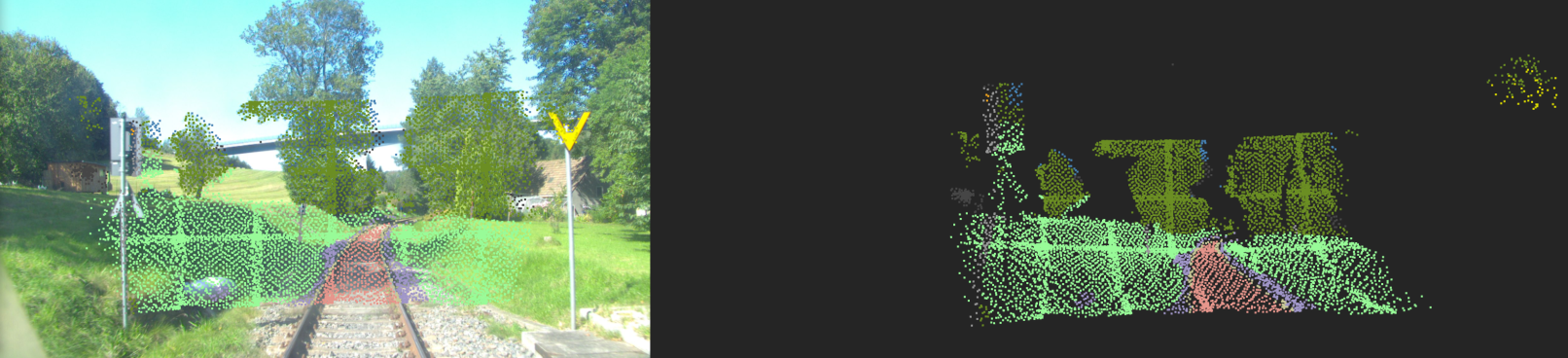} 
   \caption[prelabelfllt]{Prelabeled scene with wrongly annotated labels, e.g. sky at tree edges and background classes on traffic sign}
  \label{fig:prelabelfllt}
\end{figure*}

Large annotated datasets are essential for training neural networks in environmental perception. A dataset should include relevant and typical environmental conditions, scenarios, objects, classes, and corresponding labels for the rail environment. These include railway-specific classes, e.g., trackbeds, rail tracks, trains, or signals. Furthermore, data quality and quantity are crucial, requiring efficient annotation with minimal manual interaction.

For the creation of a dataset, we propose a semi-automated intelligent data annotation approach. Fraunhofer FOKUS develops the labeling tool FLLT.AI \cite{fllt} for 3D LiDAR data, which we used to label our rail dataset. FLLT.AI enables the visualization of image data in 2D, the annotation of point clouds in 3D, and the mapping and transfer of labels between 2D and 3D. The transformation allows for an easier manual annotation or correction of labels in 3D, as well as for automated labeling and creation of labels in the 2D image data with a following transference of labels into the LiDAR frame.

To reduce manual labeling effort, we use FLLT.AI with a trained neural network for semantic segmentation in camera images to label each pixel automatically in 2D. These labels are projected to corresponding LiDAR points, yielding automatically annotated LiDAR data. However, as the transformation is imperfect, these labels still need to be manually checked and partly corrected afterwards. Corrections are especially required for transitions between foreground and background, edges of objects, thin or small objects, or very distant objects. Some misclassifications result from the trained image segmentation network, particularly if the image contains overexposure, underexposure, or blurriness. Additionally, certain classes commonly classified in images, such as background or sky, do technically not exist in labeled LiDAR data. An automatically annotated example is shown in figure \ref{fig:prelabelfllt}.

For this work, railway-specific aspects are added to FLLT.AI. We include the neural network DeepLabV3+ \cite{deeplabv3} trained on RailSem19 \cite{railsem19} to prelabel railway-specific objects and the definition of rail-specific classes. The classes particularly for the rail environment are vehicles on track (e.g., trains), rail track, trackbed, and signals in addition to common classes such as person, construction, poles, vegetation, or terrain.

The result of this automated prelabeling is a total of 602 imperfect, coarsely labeled scans. By manual annotation and correction, a subset of 110 fine and precisely labeled scans are labeled afterwards. Additionally, we label an additional test dataset of 25 scans for evaluation. The labeled classes and their distribution in the test data are shown in section \ref{sec:evaluation} in table \ref{tab:classscore}. Even with this small size of the dataset, it is shown how the data annotation approach can be used to create a railway LiDAR dataset in an effective way.

\section{Training}
\label{sec:training}

\subsection{Networks selected for semantic segmentation}
\label{sec:networks}

\paragraph*{\textbf{DeepLabV3+}}

For the 2D semantic segmentation prelabeling in the image domain in FLLT.AI, the DeepLabV3+ \cite{deeplabv3} network is used.
DeepLabV3+ is based on an encoder and decoder architecture and uses so-called atrous or dilation convolutions and feature pyramids.
Dilation convolutions extend sparse filter kernels with specified gaps to capture a wider range and context without increasing the number of parameters or computational complexity. In addition, the feature pyramid network part achieves a good combination of different resolutions, scales, and features. The features of the encoder part are transferred or merged into the corresponding layers. These approaches achieve a robust understanding of spatial relationships and features leading to highly-performing semantic segmentation.

\begin{figure*}[t]
  \includegraphics[width=0.98\textwidth]{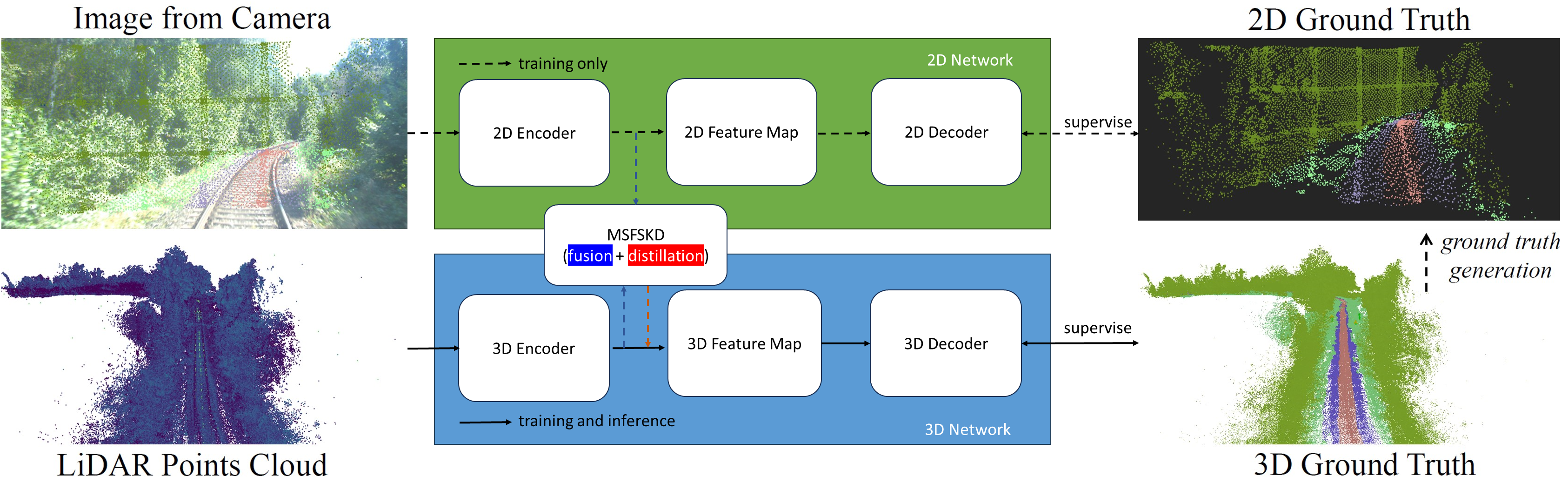} 
   \caption[2dpass]{2DPass network architecture and data pipeline adapted from Yan et al. \cite{2dpass}}
  \label{fig:2dpass}
\end{figure*}

\paragraph*{\textbf{2DPass: 2D Priors Assisted Semantic Segmentation on LiDAR Point Clouds}}
The 2DPass network \cite{2dpass} was used for the 3D semantic segmentation task in LiDAR point clouds. At the time of publication, 2DPass is one of the state-of-the-art networks in the SemanticKITTI benchmark \cite{behley2019semantickitti} with a 72.89\% mIoU. The network is real-time capable, i.e., 62 ms computation time stated by the authors, and thus enables inference within the 100 ms acquisition time of each lidar scan.
2DPass can be trained using 2D image data and 3D point clouds in parallel and jointly, with labels only required for the 3D point clouds, which are created as described above. The architecture of 2DPass combines a 2D and a 3D network path. In the training process, the 3D ground truth labels are first projected into the 2D image domain, thus generating labels for the 2D network. In this process, the 2D data, such as color information, is used as additional prior information for the 3D network. In addition, different scales and crops of the image and LiDAR data are used to leverage additional surrounding information and context in training, which is referred to as multi-scale fusion-to-single knowledge distillation (MSFSKD) as shown in figure \ref{fig:2dpass}. Inference with the trained network is possible with only the 3D point clouds, which enables a robust and performant semantic segmentation of the scene in challenging environmental conditions, which is crucial for safety-relevant rail applications.


\subsection{Network training}
\label{sec:networktraining}
For the training of the 2DPass network, the previously created dataset is used, while 25 scans are additionally annotated and used as a test dataset for final evaluation and scores reported in section \ref{sec:evaluation}. Applying transfer learning based on weights pretrained with the SemanticKITTI dataset \cite{behley2019semantickitti}, the coarsely labeled dataset with 602 scans is used for a first training iteration. As some points in 3D are incorrectly labeled automatically, the network is unable to learn a clear representation and correct interpretation yet. Erroneous labels in the training data incorrectly punish the network in some cases, even for correctly predicted labels. The classification of small or thin objects, such as poles, signs, or persons, is not correctly learned, as the labels are incorrect due to the misalignment in the 2D-3D transformation.

\subsection{Active learning}
\label{sec:active_learning}

To improve performance and robustness with as little manual effort as possible, active learning is used. This method selects additional data to retrain the network to improve the network performance as efficiently as possible with only a minimal ammount of additional training data as proposed by \cite{wang2016cost} and \cite{sener2017active}. The approach leverages predictions of the currently trained network to identify and query difficult or confusing examples or edge cases that are classified poorly. Based on statistical parameters, examples from the unlabeled data are selected for manual annotation and are added to the training dataset. We select scans based on diversity sampling and uncertainty sampling as proposed by R. Monarch \cite{monarch2021human}.

\paragraph*{\textbf{Diversity sampling}}

Diversity sampling is used to identify scans, where the average entropy $\overline{H}(scan)$ as a measure of the diversity of a scan with regard to the predicted class labels is the highest. We quantify the amount of diversity in a scan based on the entropy of the class predictions. The entropy per point $H(p)$ is calculated by the softmax probabilities $\text{softmax}(p)$ per point using the class confidences $P_{p}(y|x)$.

\begin{equation}
\begin{split}
&\overline{H}(scan)=\frac{1}{n} \sum_{p=1}^{n} H(p) \\
\text{with} \\
&H(p)=\frac{-\sum_{classes}^{}\text{softmax}(p) log_{2} (\text{softmax}(p))}{log_{2}(N_{classes})}
\end{split}
\end{equation}

                
                

\paragraph*{\textbf{Uncertainty sampling}}

Uncertainty sampling is used to identify scans with the most uncertain predicted classes, e.g., the average of the softmax score per predicted class for each point. To calculate the uncertainty of the predicted class per single point $U_{max}(p)$, the softmax probabilities are distributed within the unit interval $[0,1]$, with 1 defined as the most uncertain score. The uncertainty score per scan is calculated as the mean uncertainty of all points:
\begin{equation}
\begin{split}
&\overline{U(scan)} = \frac{1}{n} \sum_{p=1}^{n} U_{max}(p) \\
\text{with} \\
&U_{max}(p) = (1 - \text{argmax}(\text{softmax}(p)))
\end{split}
\end{equation}

                

\paragraph*{\textbf{Selecting scans}}

Finally, all scans of the remaining, unannotated dataset are selected by their entropy and uncertainty score through inference with the currently trained network. The first $n = 10$ scans are selected for active learning based on the minimal sum from the rank $R$ of both scores in each learning iteration:


\begin{equation}
R(scan) = R_{H}(scan) + R_{U}(scan)
\end{equation}


\begin{table*}[tb!]
\centering
\begin{tabular}{llll}
\hline
\rowcolor[HTML]{C0C0C0} 
\multicolumn{1}{l|}{\cellcolor[HTML]{C0C0C0}}	& \multicolumn{1}{l|}{\cellcolor[HTML]{C0C0C0}\textbf{Coarse prelabeled dataset}}	& \multicolumn{1}{l|}{\cellcolor[HTML]{C0C0C0}\textbf{Manually corrected dataset}}	& \textbf{Actively selected scans}	\\ \hline
\multicolumn{1}{l|}{\textbf{Scans}}		& \multicolumn{1}{l|}{602}		& \multicolumn{1}{l|}{52}		& 58 (+ 52) = 110	\\ \hline
\multicolumn{1}{l|}{\textbf{mean IoU}}		& \multicolumn{1}{l|}{\textbf{36.59\%}}	& \multicolumn{1}{l|}{\textbf{46.73\%}}	& \textbf{71.48\%}	\\ \hline
\multicolumn{1}{l|}{\textbf{fwIoU}}		& \multicolumn{1}{l|}{58.34\%}		& \multicolumn{1}{l|}{60.59\%}		& 81.20\%		\\ \hline
\multicolumn{1}{l|}{\textbf{Class (frequency weight)}} &	\multicolumn{3}{l}{\textbf{IoU per class}}											\\ \hline
\multicolumn{1}{l|}{\textbf{\begin{tabular}[c]{@{}l@{}}ON\_TRACKS (10.44\%):\\ PERSON (1.28\%):\\ RAIL\_TRACK (9.55\%):\\ TRACKBED (7.71\%):\\ CONSTRUCTION (6.04\%):\\ POLE (0.48\%):\\ SIGN (0.58\%):\\ VEGETATION (40.73\%):\\ TERRAIN (23.18\%):\end{tabular}}} & 
\multicolumn{1}{l|}{\begin{tabular}[c]{@{}l@{}} 67.29\%\\ 8.13\%\\ 60.02\%\\ 25.24\%\\ 60.08\%\\ 0.00\%\\ 0.00\%\\ 84.00\%\\ 24.52\%\end{tabular}} & 
\multicolumn{1}{l|}{\begin{tabular}[c]{@{}l@{}}	83.91\%\\ 79.80\%\\ 51.45\%\\ 27.85\%\\ 48.30\%\\ 16.52\%\\ 0.00\%\\ 83.21\%\\ 29.56\%\end{tabular}} & 
\begin{tabular}[c]{@{}l@{}}			88.38\%\\ 87.56\%\\ 58.04\%\\ 50.96\%\\ 76.98\%\\ 75.10\%\\ 33.92\%\\ 92.41\%\\ 79.93\%\end{tabular} \\ \hline \end{tabular}

\caption{Comparision of validation mIoU and fwIoU scores per class between training iterations}
\label{tab:classscore}
\end{table*}

\section{Evaluation}
\label{sec:evaluation}

As shown in table \ref{tab:classscore}, we achieve state-of-the-art performance scores for semantic segmentation and show the application and transfer to the railway environment.
The performance of the semantic segmentation is measured by the effectiveness of a model in correctly classifying each point in a scan into the correct predefined class label. Intersection over Union (IoU) is a metric to measures this overlap between the prediction and the ground truth class per point, calculated as the ratio of the intersection to the union of the predicted and true class. We use the mean IoU (mIoU) and the frequency-weighted IoU (fwIoU) with the weights per class reported in brackets for our test dataset.

In addition to the mIoU, which weights all 9 classes equally, the fwIoU is calculated based on class weights determined by their relative frequency of occurrence, giving more weight proportionally to classes that appear more frequently in the dataset. After the initial training with 602 scans, leveraging transfer learning based on weights pretrained with the SemanticKITTI dataset \cite{behley2019semantickitti}, we reached a mIoU of 36.59\% and a fwIoU of 58.34\%. This shows the need for the inspection and partial correction of the automatically generated and transferred labels as explained in section \ref{sec:labeling}.

Retraining with 52 manually corrected and precisely labeled scans, we achieved a mIoU of 46.73\% and a fwIoU of 60.59\%. This corresponds to an optimization of 27.71\% for the mIoU and 3.86\% for the fwIoU compared to the automatically labeled data set. Object segmentation in the foreground, especially for persons or poles, improved greatly. This is due to the fact, that most issues in the automatically labeled data were caused by imperfect calibration. Therefore, small objects in the foreground were wrongly labeled with adjacent classes from the background. Counterintuitively, the classification performance for some environmental classes, which are present mostly in the background, such as rail tracks or constructions, is even slightly decreased. Those over-represented classes were not affected as much by the imperfect coarse labeling. As the mIoU weights all classes equally, the mIoU improves significantly, benefiting especially from the big improvements in small object classes, while the fwIoU is only slightly improved as a result.

Finally, by applying active learning and performing multiple training iterations with ten additional selected scans each time, we achieve significantly improved segmentation scores of 71.48\% mIoU and 81.20\% fwIoU. This corresponds to a further optimization of 52.96\% for the mIoU and 34.02\% for the fwIoU compared to the corrected dataset and a total optimization of 95.35\% for the mIoU and 39.18\% for the fwIoU compared to the automatically generated data set. 

Especially rare and under-represented classes with only a few labeled points, such as poles, signs, or persons are further improved greatly by manual labeling and active learning. This contributes particularly to the improvement of the mIoU, as all classes are weighted equally for this score.

\begin{figure*}[!htb]
 \centering
  \includegraphics[width=0.98\textwidth, trim=0 20 0 40, clip]{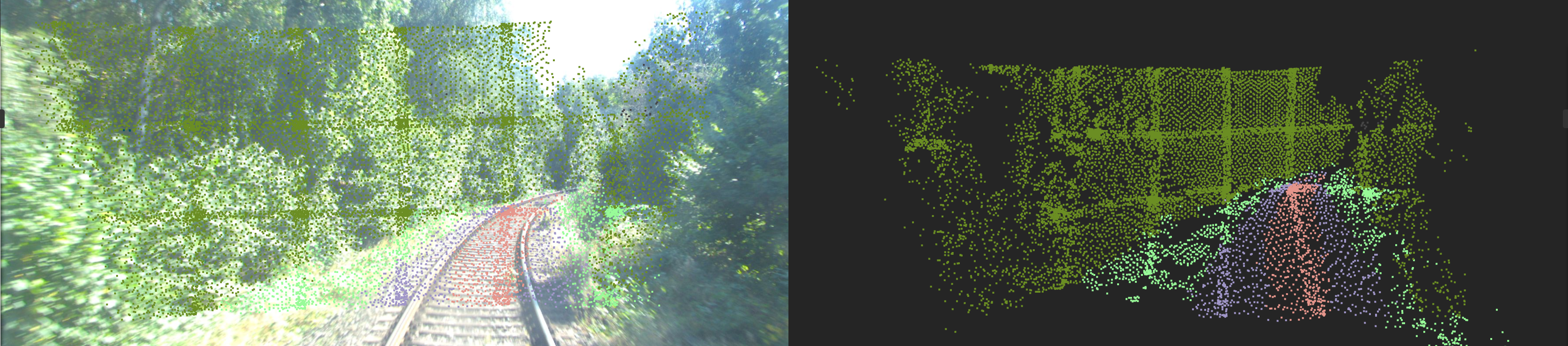} 
   \caption[inference1]{Inference example for rail track, trackbed, vegetation, and terrain}
  \label{fig:inference1}
\end{figure*}

\begin{figure*}[!htb]
 \centering
  \includegraphics[width=0.98\textwidth, trim=0 20 0 40, clip]{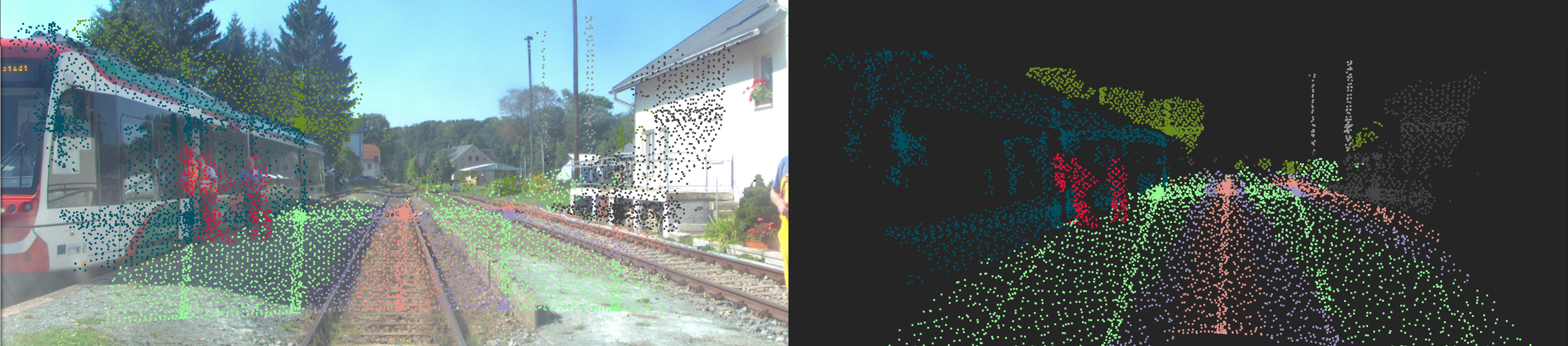} 
   \caption[inference2]{Inference example for persons moving and a train in a static scene}
  \label{fig:inference2}
\end{figure*}

\begin{figure*}[!htb]
 \centering
  \includegraphics[width=0.98\textwidth, trim=0 20 0 40, clip]{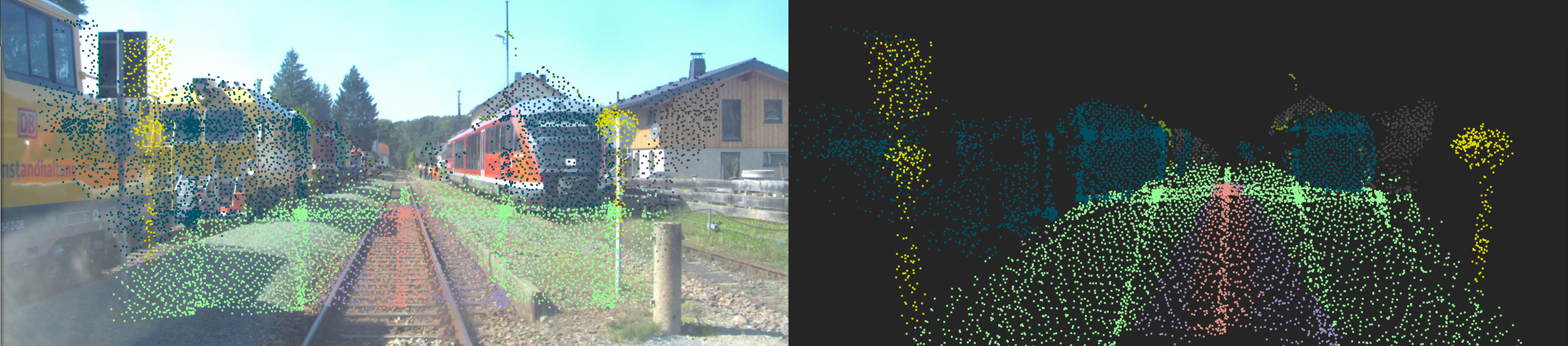} 
   \caption[inference3]{Inference example for static objects including trains, signals, and construction}
  \label{fig:inference3}
\end{figure*}

We demonstrate the successful training of the network qualitatively in exemplary scans shown in the figures \ref{fig:inference1}, \ref{fig:inference2} and \ref{fig:inference3}. Rail track, trackbed, vegetation, and terrain are successfully segmented as shown in figure \ref{fig:inference1}.
The different segments and areas of the rail track are correctly classified. The segmentation also performs track detection and ego lane detection, which are very important for fully automated trains, e.g., to detect obstacles on track and free space. The classification of adjacent areas, such as terrain or vegetation, facilitates additional use cases, such as vegetation or infrastructure monitoring.

In the second example in figure \ref{fig:inference2}, the robust segmentation and classification of static objects, such as persons and a train, is shown. This enables object detection, object tracking, or movement estimation, which are required for use cases like emergency braking systems or behavior prediction when entering into stations or to determine level crossing clearance.

In figure \ref{fig:inference3}, the segmentation and correct classification of multiple objects and structures or landmarks, such as trains, poles, signals, and constructions are shown. This can be used for signal detection, dynamic and static object detection, or precise localization with landmarks.

\section{Summary and future work}
\label{sec:summary}

Overall, we achieve a robust and accurate state-of-the-art 3D semantic segmentation with only a small amount of data. Key components of the automotive perception are successfully applied to the rail environment, thus showcasing and enabling future use cases for scene understanding for automated trains.
Our work involves full environmental segmentation and classification of rail-specific classes, domain adaption and transfer learning from automotive perception, and proposing a semi-automated annotation pipeline. Despite a small dataset, we achieve sufficient performance by incorporating camera images, using a-priori knowledge by fusion and distillation of 2D and 3D information, and applying active learning, thus reaching a mIoU of 71.48\% and a fwIou of 81.20\%.
We demonstrate an inference of the 3D semantic segmentation relying on LiDAR data only. The use of a LiDAR is especially important for the application in safety-critical domains. LiDAR is robust and largely independent of weather, daytime, or environmental conditions. Additionally, LiDAR provides long-range and free-space detection. This is very beneficial, considering that trains run at higher speeds and have significantly longer braking distances than automated cars.
Despite the challenges posed by limited datasets for semantic segmentation of environmental classes in rail, and no other published method or public dataset for this task, and the small amount of our data, semantic segmentation is successfully demonstrated as a proof of concept. Quantitative comparison with other approaches is currently not possible. However, the achieved mIoU of 71.48\% is comparable to the reported mIoU of 72.89\% for 2DPass in the semantic Kitti dataset \cite{2dpass}. The used data cannot be made public to close this gap, as the recorded data is non-public. In the highly regulated and safety-relevant railway sector, own data cannot be recorded on private and critical infrastructure. Additional work should focus on recording a public dataset to cover different environmental conditions and additional scenarios. In contrast to automotive, this severely limits the progress and contributions. Our method could be applied to the recorded data.

\newpage
\twocolumn[
\begin{center}
\addcontentsline{toc}{section}{References}
\end{center}
]
\bibliography{paper}{}
\bibliographystyle{IEEEtran}
\end{document}